%% file: main.tex
\documentclass[10pt,twocolumn,letterpaper]{article}

\usepackage{microtype}
\usepackage{iccv}
\usepackage{tikz}
\usepackage{pgfplots}
\usepackage{times}
\usepackage{mathptmx}
\usepackage{graphicx}
\usepackage{amsmath}
\usepackage{amssymb}
\usepackage{balance}
\usepackage[pagebackref=true,breaklinks=true,letterpaper=true,colorlinks,bookmarks=false]{hyperref}
\usepackage{bold-extra}
\usepackage{dsfont}
\usepackage{amsmath}
\usepackage{amssymb}
\usepackage{inconsolata}
\usepackage{amsthm}
\usepackage{float}
\usepackage{booktabs}
\usepackage{algorithm}
\usepackage{algpseudocode}
\usepackage{subcaption}
\usepackage[accsupp]{axessibility}

\pgfplotsset{compat=1.17}

\DeclareMathAlphabet{\mathcal}{OMS}{cmsy}{m}{n}

\input{colors}

\iccvfinalcopy 

\ificcvfinal\fi

\newcommand{\lab}[1]{\textquotesingle{#1}\textquotesingle}

\newcommand{\tool}[1]{\textsc{#1}\xspace}

\newtheorem{definition}{Definition}
\newtheorem{observation}{Observation}
\newtheorem{corollary}{Corollary}

\newcommand{\deepcover}{{\sc d}eep{\sc c}over\xspace}
\newcommand{\deepcoverSBFL}{{\sc d}{\sc c}-{\sc sbfl}\xspace}
\newcommand{\cet}{{\sc d}{\sc c}-{\sc c}ausal\xspace}
\newcommand{\gradcam}{{\sc g}rad-{\sc cam}\xspace}
\newcommand{\lime}{{\sc lime}\xspace}
\newcommand{\shap}{{\sc shap}\xspace}
\newcommand{\rise}{{\sc rise}\xspace}
\newcommand{\extremal}{{\sc e}xtremal\xspace}

\newcommand{\rap}{{\sc rap}\xspace}
\newcommand{\lrp}{{\sc lrp}\xspace}
\newcommand{\gbp}{{\sc gbp}\xspace}
\newcommand{\ig}{{\sc ig}\xspace}

\newcommand{\commentout}[1]{}

\newtheorem{lemma}{Lemma}

\begin{document}

\title{Explanations for Occluded Images}

\author{
Hana Chockler\\
causaLens and \\
King's College London\\
{\tt\small hana.chockler@kcl.ac.uk}
\and
Daniel Kroening\thanks{The work reported in this paper
was done prior to joining Amazon.}\\
Amazon.com, Inc.\\
{\tt\small daniel.kroening@gmail.com}
\and
Youcheng Sun\\
Queen's University Belfast\\
{\tt\small youcheng.sun@qub.ac.uk}
}

\makeatletter
\g@addto@macro\@maketitle{
\vspace{-.5cm}
\begin{figure}[H]
\setlength{\linewidth}{\textwidth}
\setlength{\hsize}{\textwidth}
\centering
  \centering
  \subfloat[Input]{
    \includegraphics[width=0.133\linewidth]{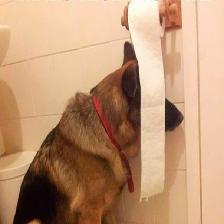}
  }
  \subfloat[\bfseries{\scshape{dc-c}}ausal]{
    \includegraphics[width=0.133\linewidth]{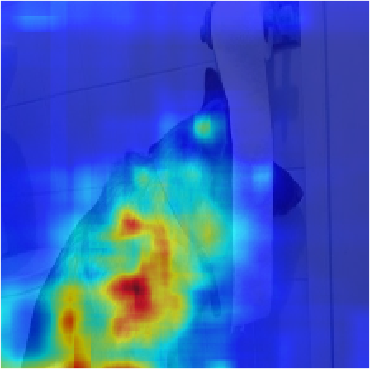}
  }
  \subfloat[\deepcoverSBFL]{
    \includegraphics[width=0.133\linewidth]{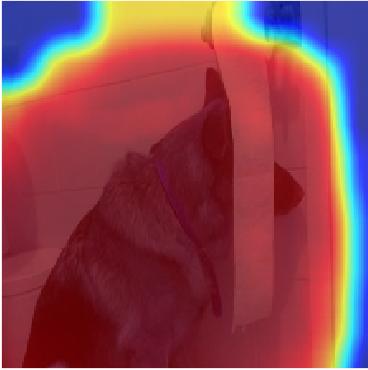}
  }
  \subfloat[\extremal]{
    \includegraphics[width=0.133\linewidth]{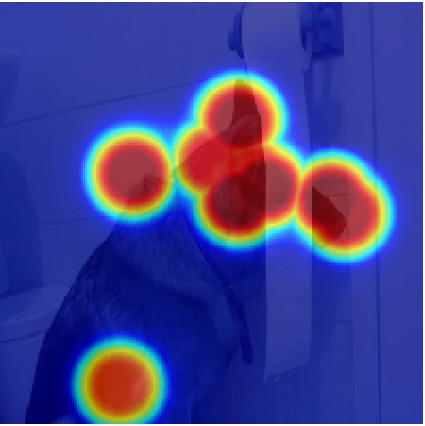}
  }
  \subfloat[\rise]{
    \includegraphics[width=0.133\linewidth]{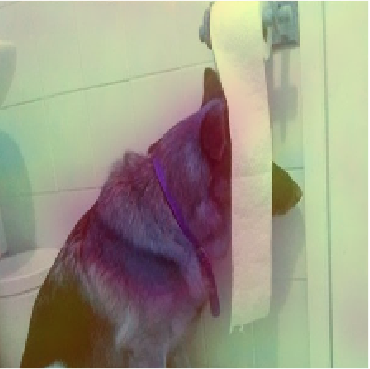}
  }
  \subfloat[\rap]{
    \includegraphics[width=0.133\linewidth]{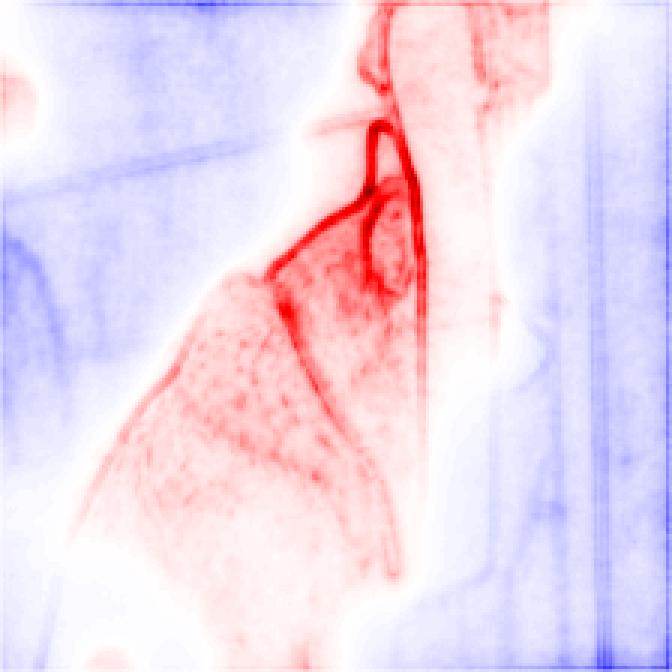}
  }
  \subfloat[\lrp]{
    \includegraphics[width=0.133\linewidth]{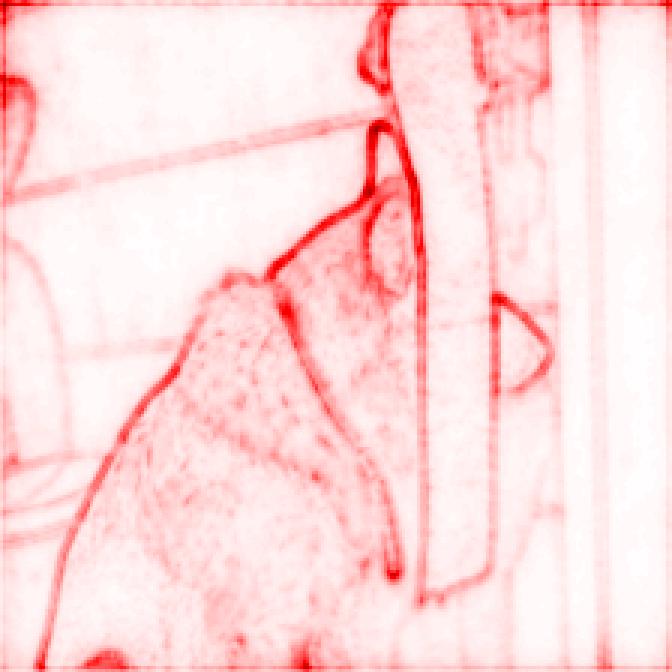}
  }
\caption{Explanations for the classification as \lab{German shepherd dog}~(a). \cet is the tool presented here.}
\label{fig:accept-figure}
\end{figure}
}

\maketitle
\thispagestyle{empty}


\begin{abstract}
Existing algorithms for explaining the output of image classifiers
perform poorly on inputs where the object of interest is partially occluded.
We present a novel, black-box algorithm for computing
explanations that uses a principled approach based on
causal theory. 
We have implemented the method in the \tool{DeepCover} tool.
We obtain explanations that are much more accurate than those generated by
the existing explanation tools on images with occlusions and observe
a level of performance comparable to the state of the art
when explaining images without occlusions.
\end{abstract}

\input{introduction.tex}

\input{related}
\input{causal_theory}
\input{algorithm}

\input{evaluation}

\input{conclusions}

\balance
\newpage
{\small
\bibliographystyle{ieee_fullname}
\bibliography{all}
\balance
}

\end{document}

%% file: colors.tex
\definecolor{tabutter}{rgb}{0.98824, 0.91373, 0.30980}          
\definecolor{ta2butter}{rgb}{0.92941, 0.83137, 0}               
\definecolor{ta3butter}{rgb}{0.76863, 0.62745, 0}               

\definecolor{taorange}{rgb}{0.98824, 0.68627, 0.24314}          
\definecolor{ta2orange}{rgb}{0.96078, 0.47451, 0}               
\definecolor{ta3orange}{rgb}{0.80784, 0.36078, 0}               

\definecolor{tachocolate}{rgb}{0.91373, 0.72549, 0.43137}       
\definecolor{ta2chocolate}{rgb}{0.75686, 0.49020, 0.066667}     
\definecolor{ta3chocolate}{rgb}{0.56078, 0.34902, 0.0078431}    

\definecolor{tachameleon}{rgb}{0.54118, 0.88627, 0.20392}       
\definecolor{ta2chameleon}{rgb}{0.45098, 0.82353, 0.086275}     
\definecolor{ta3chameleon}{rgb}{0.30588, 0.60392, 0.023529}     

\definecolor{taskyblue}{rgb}{0.44706, 0.56078, 0.81176}         
\definecolor{ta2skyblue}{rgb}{0.20392, 0.39608, 0.64314}        
\definecolor{ta3skyblue}{rgb}{0.12549, 0.29020, 0.52941}        

\definecolor{taplum}{rgb}{0.67843, 0.49804, 0.65882}            
\definecolor{ta2plum}{rgb}{0.45882, 0.31373, 0.48235}           
\definecolor{ta3plum}{rgb}{0.36078, 0.20784, 0.4}               

\definecolor{tascarletred}{rgb}{0.93725, 0.16078, 0.16078}      
\definecolor{ta2scarletred}{rgb}{0.8, 0, 0}                     
\definecolor{ta3scarletred}{rgb}{0.64314, 0, 0}                 

\definecolor{taaluminium}{rgb}{0.93333, 0.93333, 0.92549}       
\definecolor{ta2aluminium}{rgb}{0.82745, 0.84314, 0.81176}      
\definecolor{ta3aluminium}{rgb}{0.72941, 0.74118, 0.71373}      

\definecolor{tagray}{rgb}{0.53333, 0.54118, 0.52157}            
\definecolor{ta2gray}{rgb}{0.33333, 0.34118, 0.32549}           
\definecolor{ta3gray}{rgb}{0.18039, 0.20392, 0.21176}           

%% file: introduction.tex
\section{Introduction}
\label{sec:introduction}

Deep neural networks (DNNs) are now a primary building block of many computer
vision systems. DNNs are complex non-linear functions with algorithmically
generated (and not engineered) coefficients. In contrast to traditionally
engineered image processing pipelines it is difficult to retrace how the
pixel data are interpreted by the layers of the DNN.
This ``black box" nature of DNNs creates demand for techniques that explain
why a particular input yields the output that is observed. 

An explanation of an output of an automated procedure is essential in many areas, including verification,
planning, diagnosis and the like. A good explanation can increase a user's confidence
in the result. Explanations are also useful for determining whether there is a fault in the automated procedure:
if the explanation does not make sense, it may indicate that the procedure is faulty. 
It is less clear how to define what a \emph{good} explanation is. There have been a number of definitions of explanations over the years in various domains of computer science~\cite{CH97,Gar88,Pea88}, philosophy~\cite{Hem65} and statistics~\cite{Sal89}.
The recent increase in the number of machine learning applications and the advances in deep learning led to the need
for \textit{explainable AI}, which is advocated, among others, by DARPA~\cite{DARPA} to promote understanding, trust, and adoption of future autonomous systems based on learning algorithms (and, in particular, image classification DNNs).
DARPA provides a list of questions that a good explanation should answer and an epistemic state of the user after receiving a good explanation. The description of this epistemic state boils down to \emph{adding useful information} about the output of the algorithm and \emph{increasing trust} of the user in the algorithm.
 
Explanations for the results of image classifiers are typically based on or 
are given in
the form of a \emph{ranking} of the pixels, which is a numerical
measure of importance: the higher the score, the more important the pixel
is for the DNN's classification outcome. 

A user-friendly explanation can then be a subset of highest-ranked pixels that
is sufficient for the original classification outcome.
Given an image that features an
object, good algorithms are able to generate
rankings that identify that object with a high accuracy. 
Another typical proxy for the quality of a ranking is how many of the
high-ranked pixels (constituting an explanation) need to be masked before the classification
generated by the DNN changes. Good explanations require very little
masking. 

\commentout{Leading tools
in the area include (in order of publication)
\lime~\cite{lime}, 
\shap~\cite{datta2016algorithmic}, 
\gradcam~\cite{CAM}, 
\rise~\cite{petsiuk2018rise}, 
\extremal~\cite{fong2019understanding} 
and
\deepcover~\cite{sun2020explaining}. 
}


In the absence of further analysis, the space of possible orderings (and hence rankings)
is exponential in the size of the image, and the brute-force approach is
therefore impractical. Moreover, as we argue in this paper, the problem is NP-complete.
It is therefore expected that explanation-generating algorithms approximate the solution
using heuristics tuned for image classification.
%
%
This assumption is entirely appropriate in many use cases,
and in particular, works very well on the benchmark sets that are used
in the area: the existing work has been evaluated using the ImageNet
dataset and ImageNet has been curated so are all objects clearly
visible. Consequently, the explanations that are generated are usually contiguous.

We argue that there is a use-case for explanations of the results of image
classifiers for images where the trigger for the result is \emph{not}
contiguous. Obvious exemplars are images with partial occlusion, say
by a person walking into a scene or simply by dirt on your camera lens.
To quantify the quality of the explanations for such images objectively,
we introduce a new image dataset we call \emph{Photo Bombing},
in which we obscure 
ImageNet photos by masking parts of the object.
The difference between the modified image and the original one is the
ground truth for the ``photobomber'', and a good explanation has no
overlap with it. 

We observe that the existing methods for explaining the outcome of image
classifiers perform poorly on such inputs. To address this problem,
we introduce a new algorithm that is grounded in causal theory. Our algorithm
is iterative and highly parallelizable and delivers
significantly better accuracy on an existing dataset with partial occlusion
and on our own photo bombing data set. The tool, the new benchmark set,
and the full set of results are available at
\url{https://www.cprover.org/deepcover/}.

%% file: related.tex
\section{Related Work}
\label{sec:related}

There is a large body of work on explaining image classifiers. The existing approaches can be largely grouped into two categories: propagation and perturbation. 

Propagation-based explanation methods are often regarded as more efficient. They back-propagate a model's decision to the input layer to determine the weight of each input feature for the decision.
\gradcam~\cite{CAM} only needs one backward pass and propagates the class-specific gradient into the final convolutional layer of a DNN to coarsely highlight important regions of an input image. 
Guided Back-Propagation (\gbp) \cite{springenberg2015striving} computes the single and average partial derivatives of the output to attribute the prediction of a DNN. Integrated Gradients (\ig) \cite{sundararajan2017axiomatic} further uses two axioms called sensitivity and implementation invariance for the problem of how to attribute the classification by a deep network to its input features. Layer-wise relevance propagation (\lrp)~\cite{bach2015pixel} is defined by a set of constraints on the layers and it assumes that the classifier can be decomposed into several layers of computation.
In~\cite{shrikumar2017learning}, the activation level of each neuron is compared with some reference point, and its contribution score for the final output is assigned according to the difference. 
\rap (Relative Attributing Propagation)~\cite{nam2020relative} attributes a positive and negative relevance
to each neuron, according to its relative influence among the neurons. 
SHAP (SHapley Additive exPlanations)~\cite{lundberg2017unified} goes beyond propagation and identifies inputs that are similar to the input for which the output is to be explained, and ranks the features of the input according to their difference. A key advantage of SHAP is that it does not require counterfactuals.

In contrast to propagation-based explanation methods, perturbation-based explanation approaches explore the input space directly
in search for an explanation. The exploration/search often requires a large
number of inference passes, which incurs significant computational cost when
compared to propagation methods. Many sampling methods have been proposed, but
most of are based on random search or heuristics and lack rigor.

Given a particular input, \lime~\cite{lime} samples the the neighborhood of
this input and creates a linear model to approximate the system's local behavior;
owing to the high computational cost of this approach, the ranking uses
super-pixels instead of individual pixels. In~\cite{datta2016algorithmic},
the natural distribution of the input is replaced by a user-defined distribution
and the Shapley Value method is used to analyze combinations of input features
and to rank their importance.  In~\cite{chen2018learning}, the importance of input
features is estimated by measuring the the flow of information between inputs and
outputs. Both the Shapley Value and the information-theoretic approaches are
computationally expensive.  In \rise~\cite{petsiuk2018rise}, the importance of a pixel
is computed as the expectation over all local perturbations conditioned on the event
that the pixel is observed. The concept of ``extreme perturbations'' has been
introduced to improve the perturbation analysis by the \extremal
algorithm~\cite{fong2019understanding}. More recently, spectrum-based fault
localisation (SBFL) has been applied to explaining image classifiers. The technique
has been implemented in the tool \deepcover~\cite{sun2020explaining}, and we refer
to it as \deepcoverSBFL. It outperforms the other tools when explaining images without occlusions.
Both \rise and \deepcoverSBFL mask input pixels randomly. By contrast, \extremal uses an area constraint to optimize the perturbation. Like our new method, \deepcoverSBFL constructs explanations greedily from a ranked list of pixels. The ranking, however, is calculated with
statistical fault localisation, and is much less precise, as we demonstrate
empirically.

The work presented in this paper is motivated by the fact that compositionality
is a fundamental aspect of human cognition~\cite{bienenstock1997compositionality, fidler2014learning}
and by the compositional computer vision work in recent years~\cite{kortylewski2020combining, wang2015unsupervised, kortylewski2020combining, zhu2019robustness, xiao2019tdapnet, zhang2018deepvoting}. While it is well known that the performance of conventional convolutional neural networks degrades when given partially occluded objects, the impact of partial occlusion on algorithms for generating explanations has not been studied before. 

The \cet method presented in this paper is a perturbation-based approach and addresses the limitations of existing black-box methods in two aspects. The feature masking in \cet uses causal reasoning that provides a guarantee (subject to the assumption). Furthermore, the \cet algorithm is highly parallelizable, which makes it ideal for large-scale computer vision problems.
As we demonstrate in Section~\ref{sec:exp0}, \cet constructs its explanations in a compositional manner and is therefore a perfect fit for compositional computer vision pipelines. 

%% file: causal_theory.tex
\section{Theoretical Foundations}

In this section we describe the theoretical foundations of our approach. 

\input{cause}

\subsection{Causes in image classification}
\label{sec:causality}
Our definition of causality for image classification follows the contingency-based approach and is derived from the definition in~\cite{HP05a} and its matching
definition of responsibility~\cite{CH04}; the variables represent the pixels of an input image. We cannot assume any dependency between the pixels of the image, hence we do not define any structural
equations on the variables. As our goal is ranking of the pixels according to their importance for the
classification, we only consider \emph{singleton causes}.

\begin{definition}[Singleton cause for image classification]\label{simple-cause}
For an image $x$ classified by the DNN as $f(x)=o$, a pixel $p_i$ of $x$ is a \emph{cause}
of $o$ iff there exists a subset $P_j$ of pixels of $x$ such that the following conditions hold:
\begin{description}
 \itemsep0em
    \item[SC1.] $p_i \not\in P_j$;
    \item[SC2.] changing the color of any subset $P'_j \subseteq P_j$ to the masking color does not change the classification;
    \item[SC3.] changing the color of $P_j$ and the color of $p_i$ to the masking color changes the classification.
\end{description}
We call such $P_j$ a \emph{witness} to the fact that $p_i$ is a cause of $x$ being classified as $o$.
\end{definition}
\begin{definition}[Simplified responsibility]~\label{simple-resp}
The \emph{degree of responsibility} $r(p_i,x,o)$ of $p_i$ for $x$
being classified as $o$ is defined as $1/(k+1)$, where $k$ is the size
of the smallest witness set $P_j$ for $p_i$. If $p_i$ is not a cause, $k$ is defined as $\infty$,
and hence $r(p_i,x,o)=0$. If changing the color of $p_i$
alone to the masking color results in a change in the classification, we have $P_j = \emptyset$, and hence
$r(p_i,x,o)=1$.
\end{definition}

\begin{lemma}\label{lemma-cause}
Definition~\ref{simple-cause} is equivalent to the definition of actual cause when all variables in the model are independent
of each other.
\end{lemma}
\begin{proof}[Proof sketch]
The definition matches the definition of a singleton cause for binary causal models, where a variable has the value $1$ if the corresponding pixel is set to its original color, and $0$ if the pixel is set to the masking color. 
The value of $\varphi$ is $1$ for the original classification (and $0$ otherwise). 
The \emph{context} assigns all variables the value $1$, corresponding to the
original image; hence $\varphi$ has the value $1$.
The minimality requirement is satisfied immediately, given that our causes are singletons. The additional condition
in~\cite{HP05a} that requires subsets $Z$ to be set to their original values is only relevant when there are dependencies
between the variables. 
\end{proof}

\begin{corollary}
The problem of detecting causes in image classification is NP-complete.
\end{corollary}
This result follows from Lemma~\ref{lemma-cause} and~\cite{EL02}.

\begin{observation}
Given an image $x$ and its classification $o$, we can calculate the degree of responsibility of each pixel $p_i$ of $x$
by directly applying Def.~\ref{simple-cause}, that is, by checking the conditions {\bf SC1}, {\bf SC2}, and {\bf SC3}
for all subsets $P_j$
of pixels of $x$ and then choosing a smallest witness subset. While there is an underlying Boolean formula that determines
the classification $o$ given the values of the pixels of $x$, we do not need to discover this formula in order to calculate
the degree of responsibility of each pixel of $x$.
\end{observation}

\subsection{Explanations in image classification}
\label{sec:explanations}

We adapt the definition of explanations by Halpern and Pearl~\cite{HP05} to our setting.
The definition in~\cite{HP05a} is derived from the definition of actual causality. Our definition is based on Def.~\ref{simple-cause}.

\begin{definition}[Explanation for image classification]\label{simple-exp}
An explanation in image classification is a minimal subset of pixels of a given input image that is sufficient for the DNN to
classify the image, where ``sufficient'' is defined as containing only this subset of pixels from the original image,
with the other pixels set to the masking color.
\end{definition}

We note that (1) the explanation cannot be too small (or empty), as a too small subset of pixels would violate the sufficiency requirement, and (2) there can be multiple explanations for a given input image.

The precise computation of an explanation in our setting is intractable, as the problem is equivalent to the earlier definition of
explanations in binary causal models, which is DP-complete~\cite{EL04} (the proof is similar to the proof of Lemma~\ref{lemma-cause}).
The brute-force approach of checking the effect of changing the color of each subset of
pixels of the input image to the masking color is exponential in the size of the image. 
Instead, we introduce a \emph{greedy compositional approach} to computing explanations. The approach is greedy because we
rank the elements in the decreasing order of responsibility for the classification and greedily add them to the explanation one
by one until the original classification is restored. This approach generates explanations that are minimal in the sense that no pixels
can be removed from them without altering the classification; however, they are not necessarily minimal in size, and hence are,
strictly speaking, an \emph{approximation} of Def.~\ref{simple-exp}.

Unfortunately, extracting approximate explanations from the ranking is still intractable, as the ranking is based on
computing the degree of responsibility of each pixels, which is NP-complete\footnote{The decision problem is NP-complete; the
corresponding function problem is FP$^{\mbox{NP}[\log{n}]}$-complete.}.

In the next section we introduce a compositional approach to computing 
the (approximate) degree of responsibility. The approach is based on the notion of a \emph{super-pixel} $P_i$, which is a subset of
pixels of a given image. Given an image $x$, we partition it into a small number of superpixels and
compute their degree of responsibility for the output of the DNN. Then, we only refine
those superpixels with a high responsibility
(exceeding a predefined threshold).
The scalability of the approach relies on the following observation, which is heuristically true in our experiments.

\begin{observation}~\label{obs:highest}
The pixels with the highest responsibility for the DNN's decision are located in super-pixels
with the highest responsibility.
\end{observation}

Intuitively, the observation holds when pixels with high responsibility do
not appear in the superpixels surrounded by other pixels with very
low responsibility for the input image classification outcome.
While this can happen in principle, we do not encouter this case
in practice owing to the continuous nature of images
(even when the explanation is non-contiguous).
This property is key to the success of our algorithm.

%% file: cause.tex

\subsection{Background on Actual Causality}

Our definitions are based on the framework of \emph{actual causality} introduced in~\cite{HP05a}.
Due to the lack of space, we do not present the framework
here in full, but instead discuss the intuition informally.
The definition of an \emph{actual cause} is based on the concept of \emph{causal models}, which consists of
the set of variables, the range of each variable, and the structural equations describing the
dependencies between the variables. Actual causes are defined with respect to a given causal model, 
a given context (an assignment to the variables of the model), and a propositional logic formula that
holds in the model in this context.

\emph{Actual causality} extends the simple counterfactual reasoning~\cite{Hume39}
by considering \emph{contingencies}, which are changes of the current setting.
Roughly speaking, a subset of variables~$X$ and their values in a given context
is an actual cause of a Boolean formula~$\varphi$ being True if there exists a change
in the values of other values that creates a counterfactual dependency
between the values of $X$ and $\varphi$ (that is, if we change the values
of variables in~$X$, $\varphi$~would be falsified). The formal definition is more
complex and requires that the dependency is not affected by changing the values
of variables not in the contingency, as well as requesting
minimality.\footnote{In~\cite{Hal15}, Halpern presents an updated definition
of causality; the version in~\cite{HP05a} is more suitable for our purposes,
as we are interested in causes consisting of a single element.}

\emph{Responsibility}, defined in~\cite{CH04}, is a quantification of causality,
attributing to each actual cause its \emph{degree of responsibility}, which is
based on the size of a smallest contingency required to create a counterfactual
dependence. Essentially, the degree of responsibility is defined as $1/(k+1)$,
where $k$ is the size of a smallest contingency. The degree of responsibility
of counterfactual causes is therefore $1$ (as $k=0$), and the degree of
responsibility of variables that have no causal influence on $\varphi$
is $0$, as $k$ is taken to be $\infty$. In general, the degree of responsibility
is always between $0$ and $1$, with higher values indicating a stronger
causal dependency.

%% file: algorithm.tex
\section{Compositional Explanations}
\label{sec:algorithm}

In this section, we present our \emph{greedy compositional explanation (CE)} algorithm.
The general idea is to calculate the responsibility of a superpixel and recursively distribute this responsibility
to all pixels within this superpixel. The CE approach in this work consists of three steps.

\begin{enumerate}
    \item Given a set of superpixels, compute the responsibility of each its superpixel (Section \ref{sec:responsibility}).
    \item Following the responsibility result in Step 1, further refine the superpixel and calculate the
    responsibility for the refined superpixels (Section \ref{sec:refine_responsibility}). 
    \item As it is insufficient to explain an input by only using one particular set of superpixels,
    multiple sets will be selected and they will be analysed independently by Step 2. Finally, all their results
    will be merged and a ranking of pixels following their responsibility will be computed, from which an explanation will be constructed (Section \ref{sec:compositional_explanation}).
\end{enumerate}
Section~\ref{sec:exp0} gives a step-by-step example to illustrate the working of the algorithm.

\subsection{Computing the responsibility of a superpixel}
\label{sec:responsibility}

Given a set of pixels $\mathcal{P}$, we use $\mathds{P}_i$ to denote a \emph{partition} of $\mathcal{P}$, that is, a 
set $\{P_{i,j}: \bigcup P_{i,j} = \mathcal{P}$ and $\forall j\not=k, P_{i,j} \cap P_{i,k}=\emptyset \}$.
The number of elements in $\mathds{P}_i$ is a parameter, denoted by $s$; in this work, we consider $s=4$.
We refer to $P_{i,j}$ as \emph{superpixels}.

For a DNN $\mathcal{N}$, an input $x$, and a partition $\mathds{P}_i$, we can generalize Def.~\ref{simple-cause}
to the set of \emph{superpixels} defined by $\mathds{P}_i$.
We denote by $r_i(P_{i,j},x,\mathcal{N}(x))$ the \emph{degree of responsibility} of
a superpixel $P_{i,j}$ for $\mathcal{N}$'s classification of $x$, given $\mathds{P}_i$.

For a partition $\mathds{P}_i$, we denote by $X_i$ the set of \emph{mutant images} obtained from $x$ by masking subsets
of $\mathds{P}_i$, and by $\tilde{X}_i$ the subset of $X_i$ that is classified as the original image $x$. Formally,
\[ \tilde{X}_i = \{ x_m : \mathcal{N}(x_m) = \mathcal{N}(x) \}. \]

We compute $r_i(P_{i,j},x,\mathcal{N}(x))$, the responsibility of each superpixel $P_{i,j}$ 
in the classification of $x$, in Alg.~\ref{algo:responsibility}.
For a superpixel $P_{i,j}$, we define the set 
\[ \tilde{X}_i^j = \{ x_m : P_{i,j} \mbox{ is not masked in } x_m \} \cap \tilde{X}_i. \]
For a mutant image $x_m$, we define $\mathit{diff_i}(x_m,x)$ as the number of superpixels in the partition $\mathds{P}_i$ that
are masked in $x_m$ (that is, the difference between $x$ and $x_m$ with respect to $\mathds{P}_i$).
For an image $y$, we denote by $y(P_{i,j})$ an image that is obtained by
masking the superpixel $P_{i,j}$ in $y$. 

The degree of responsibility of a superpixel $P_{i,j}$ is calculated by Alg.~\ref{algo:responsibility}
as a minimum difference between a mutant image and the original image over all mutant images $x_m$ that
do not mask $P_{i,j}$, are classified the same as the original image~$x$, and masking $P_{i,j}$ in $x_m$
changes the classification. 


\begin{algorithm}[!htp]
  \caption{$\mathit{responsibility}(x, \mathds{P}_{i})$}
  \label{algo:responsibility}
  \begin{flushleft}
    \textbf{INPUT:} an image $x$, a partition $\mathds{P}_i$ \\
    \textbf{OUTPUT:} a responsibility map $\mathds{P}_i \rightarrow \mathds{Q}$
  \end{flushleft}
  \begin{algorithmic}[1]
    \For{each $P_{i,j}\in \mathds{P}_i$}
        \State $k \leftarrow \min\limits_{x_m} \{ \mathit{diff}(x_m, x) \,| \,\, x_m \in \tilde{X}^j_i \}$
        \State $r_{i,j} \leftarrow \frac{1}{k+1}$
    \EndFor    
    \State \Return  $r_{i,0},\dots,r_{i,|P_i|-1}$
  \end{algorithmic}
\end{algorithm}

\subsection{Compositional refinement of the responsibility}
\label{sec:refine_responsibility}

Alg.~\ref{algo:responsibility} calculates the responsibility of each superpixel, subject to a given partition.
Then, it proceeds with only the high-responsibility superpixels. Note that in general, it is possible that all
superpixels in a given partition have the same responsibility.
Consider, for example, a situation where the explanation is right in the middle of the image, and our partition
divides the image into four quadrants. Each quadrant would be equally important for the classification, hence
we would not gain any insight into why the image was classified in that particular way. In this case, the algorithm chooses
another partition.

Our compositional algorithm (see Alg.~\ref{algo:compositonal_responsibility}) iteratively refines
the high-responsibility superpixels until a precise explanation is constructed
and recursively applies Alg.~\ref{algo:responsibility} to each refinement.

\begin{algorithm}[!htp]
  \caption{$\mathit{compositional\_responsibility}(x, \mathds{P}_i)$}
  \label{algo:compositonal_responsibility}
  \begin{flushleft}
    \textbf{INPUT:}\,\, an image $x$ and a partition $\mathds{P}_i$\\
    \textbf{OUTPUT:}\,\, a responsibility map $\mathds{P}_i \longrightarrow \mathds{Q}$
  \end{flushleft}
  \begin{algorithmic}[1]
    \State $R \leftarrow \mathit{responsibility}(x, \mathds{P}_i)$ 
    \If {$R$ meets termination condition}
        \State \Return $R$
    \EndIf

    \State $R' \leftarrow \emptyset$
    \For{each $P_{i,j} \in \mathds{P}_i$~s.t.~$R(P_{i,j})\not =0$}
        \State $R' \leftarrow R' \, \cup\, \mathit{compositional\_resposibility}(x, P_{i,j})$
    \EndFor
    \State \Return $R'$
  \end{algorithmic}
\end{algorithm}

Given a partition, Alg.~\ref{algo:compositonal_responsibility} calculates the responsibility for each
superpixel (Line 1). If the termination condition is met (Lines 2--3), the responsibility map $Q$ is updated accordingly. 
Otherwise, for each superpixel in $\mathds{P}_i$ with responsibility higher than $0$, 
we refine it and call the algorithm recursively ($0$ is a parameter that can be replaced with a sufficiently low threshold
without affecting the quality of the explanation; $\mathds{P}_{i,j}$ is an arbitrary partition of the superpixel $P_{i,j}$). 
We use $\cup$ to include these newly computed values in the returned map.
The algorithm terminates when: 
1) the superpixels in $\mathds{P}_i$ are sufficiently refined (containing only very few pixels), or
2) when all superpixels in $\mathds{P}_i$ have the same responsibility (this condition is for efficiency).

\subsection{Compositional explanation algorithm}
\label{sec:compositional_explanation}

So far, we assume one particular partition $\mathds{P}_i$, which 
Alg.~\ref{algo:compositonal_responsibility} recursively refines and calculates the corresponding
responsibilities of superpixels in each step by calling Alg.~\ref{algo:responsibility}.
We note that the choice of the initial partition can affect the values calculated
by Alg.~\ref{algo:compositonal_responsibility}, as the partition determines the set of mutants
in Alg.~\ref{algo:responsibility}.
We ameliorate the influence of the choice of any particular partition by iterating the algorithm over a set of partitions.
In Alg.~\ref{algo:compositonal_explanation}, we consider $N$ partitions and compute an average
of the degrees of responsibility induced by each of these partitions.
In the algorithm, $\mathds{P}^x$ stands for a specific partition chosen randomly from the set of partitions, and $r_p$ denotes the degree of responsibility of a pixel $p$ w.r.t.~$\mathds{P}^x$.

\begin{algorithm}[!htp]
  \caption{$\mathit{compositional\_explanation}(x)$}
  \label{algo:compositonal_explanation}
  \begin{flushleft}
    \textbf{INPUT:}\,\, an input image $x$, a parameter $N \in \mathds{N}$\\
    \textbf{OUTPUT:}\,\, an explanation $\mathcal{E}$ 
  \end{flushleft}
  \begin{algorithmic}[1]
    \State $r_p \leftarrow 0$ for all pixels $p$
    \For{$c$ in $1$ to $N$}
        \State $\mathds{P}^x\leftarrow$ sample a partition
        \State $R \leftarrow \mathit{compositional\_responsibility}(x, \mathds{P}^x)$
        \For{each $P_{i,j} \in~\mbox{domain of}~R$}
            \State $\forall p\in P_{i,j}: r_p \leftarrow r_p + \frac{R(P_{i,j})}{|P_{i,j}|}$
        \EndFor
    \EndFor
    \State $\mathit{pixel\_ranking} \leftarrow$ pixels from high $r_p$ to low
    \State $\mathcal{E}\leftarrow\emptyset$
    \For{each pixel $p_i \in \mathit{pixel\_ranking}$}
        \State $\mathcal{E}\leftarrow\mathcal{E}\cup\{p_i\}$
        \State $x^\mathit{exp}\leftarrow$ mask pixels of $x$ that are \textbf{not} in $\mathcal{E}$
        \If{$\mathcal{N}(x^\mathit{exp})=\mathcal{N}(x)$} 
          \State \Return{$\mathcal{E}$}
        \EndIf
    \EndFor
  \end{algorithmic}
\end{algorithm}

Alg.~\ref{algo:compositonal_explanation} has two parts:
ranking all pixels (Lines 1--9) and constructing the explanation (Lines 10--17). 
The algorithm ranks the pixels of the image according to their responsibility for 
the model's output. 
Each time a partition is randomly selected (Line 3), the
compositional refinement (Alg.~\ref{algo:compositonal_responsibility}) is called to refine it into
a set of fine-grained superpixels and calculate their responsibilities (Line~4). A superpixel's responsibility
is evenly distributed to all its pixels, and the pixel-level responsibility is updated accordingly for each sampled partition (Lines~5--7). After $N$ iterations, all pixels are ranked according to their responsibility $r_p$.

The remainder of Alg.~\ref{algo:compositonal_explanation} follows the method for explaining the result of an image classifier in~\cite{sun2020explaining}. That is, we construct a subset of pixels
$\mathcal{E}$ to explain $\mathcal{N}$'s output on this particular input~$x$ \emph{greedily}.
We add pixels to $\mathcal{E}$ as long as $\mathcal{N}$'s output on $\mathcal{E}$ does not
match $\mathcal{N}(x)$. This process terminates when $\mathcal{N}$'s output is the same
as on the whole image $x$. 
The set $\mathcal{E}$ is returned as an explanation. 

While we approximate the computation of an explanation in order to ensure efficiency of our approach, the algorithm is built on
solid theoretical foundations, which distinguishes it from other random or heuristic-based approaches. 
In practice, while our algorithm uses an iterative average of a greedy approximation, it yields highly accurate results. The tool website features a comparison with the exact computation of explanations for small images of 4x4 pixels, showing that \cet's explanations are optimal in more than 96\% of the cases. 
Furthermore, our approach
is simple and general, and uses the DNN as a blackbox. Finally, an important advantage of our algorithm is its high
parallelizability, as each partition can be analyzed in parallel. This opens an opportunity for further performance improvements.

%
%
%
%

\subsection{Illustrative example}
\label{sec:exp0}

\begin{figure*}
\centering
  \begin{subfigure}[b]{0.22\textwidth}
        \centering
        \includegraphics[width=\textwidth]{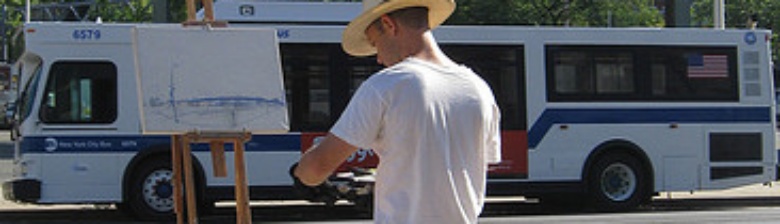}
        \caption{A partially occluded image that is classified as \lab{bus} by the compositional net~\cite{kortylewski2020compositional}}
        \label{fig:demo}
  \end{subfigure}\hspace{0.25cm}
  \begin{subfigure}[b]{0.35\textwidth}
        \centering
        \begin{tabular}{@{}c@{\hspace{0.25cm}}c}
        \includegraphics[width=0.45\linewidth]{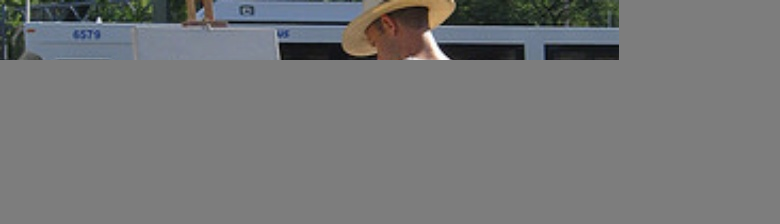} &
        \includegraphics[width=0.45\linewidth]{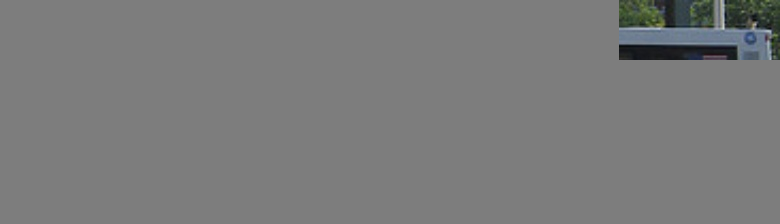}\\
        \includegraphics[width=0.45\linewidth]{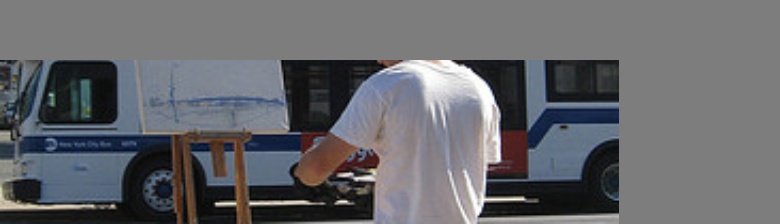} &
        \includegraphics[width=0.45\linewidth]{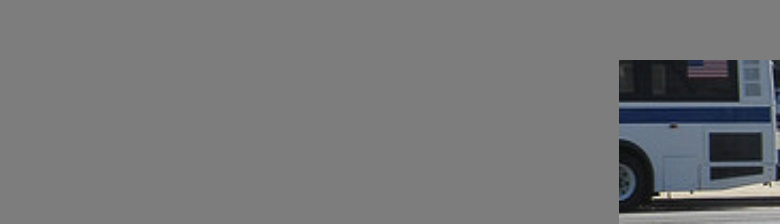}
    \end{tabular}
    \caption{The initial four superpixels chosen by \cet (Line 3 in Alg.~\ref{algo:compositonal_explanation})}
    \label{fig:demo-level0}
  \end{subfigure}\hspace{0.25cm}
  \begin{subfigure}[b]{0.35\textwidth}
        \centering
        \begin{tabular}{@{}c@{\hspace{0.25cm}}c}
        \includegraphics[width=0.45\linewidth]{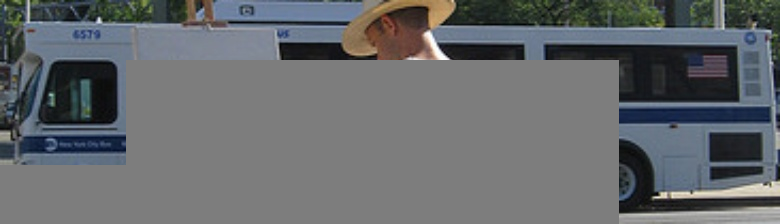} &
        \includegraphics[width=0.45\linewidth]{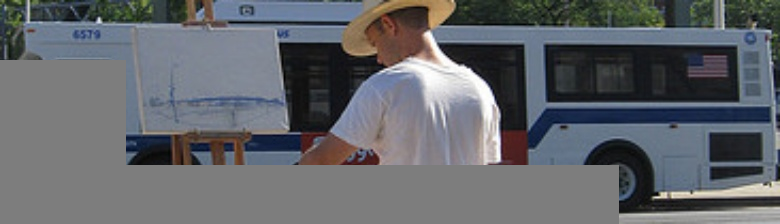}\\
        \includegraphics[width=0.45\linewidth]{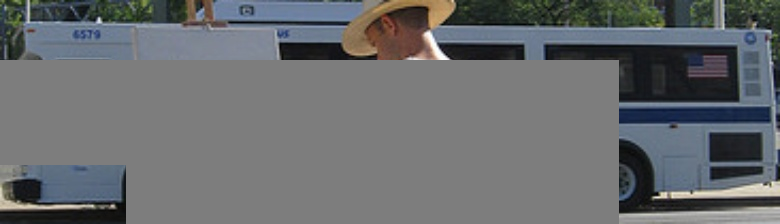} &
        \includegraphics[width=0.45\linewidth]{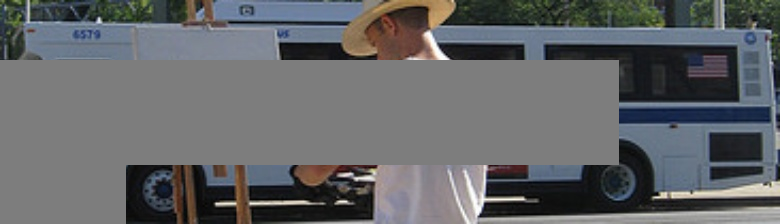} \\
    \end{tabular}
    \caption{Further refinement of the (lower left) superpixel in Figure~\ref{fig:demo-level0} (Line 7 in Alg.~\ref{algo:compositonal_responsibility})}
    \label{fig:demo-level1}
  \end{subfigure}
  \caption{An illustrative example using \cet to explain an image from the partial occlusion image data set~\cite{kortylewski2020combining, wang2015unsupervised}} 
  \label{fig:main}
\end{figure*}

\begin{figure}
  \centering
  \subfloat[$N=1$]{
    \includegraphics[width=0.23\linewidth]{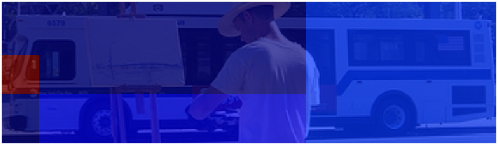}
  }
  \subfloat[$N=10$]{
    \includegraphics[width=0.23\linewidth]{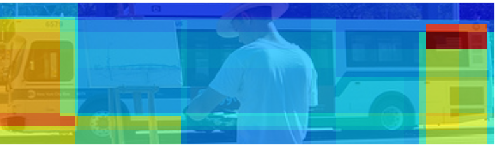}
  }
  \subfloat[$N=20$]{
    \includegraphics[width=0.23\linewidth]{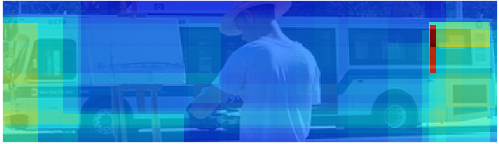}
  }
  \subfloat[$N=30$]{
    \includegraphics[width=0.23\linewidth]{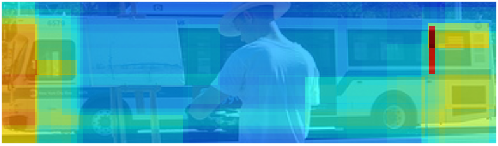}
  } \\
    \subfloat[$N=40$]{
    \includegraphics[width=0.23\linewidth]{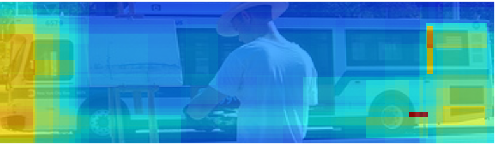}
  }
  \subfloat[$N=50$]{
    \includegraphics[width=0.23\linewidth]{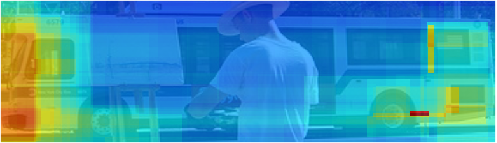}\label{bus-n50}
  }
  \subfloat[$N=70$]{
    \includegraphics[width=0.23\linewidth]{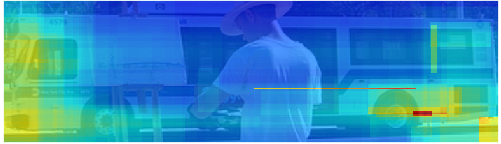}
  }
  \subfloat[$N=100$]{
    \includegraphics[width=0.23\linewidth]{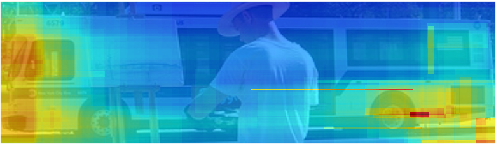}
  }
\caption{Improvement of the \cet's pixel ranking as the number of initial partitions $N$ increases (Alg.~\ref{algo:compositonal_explanation})}
\label{fig:demo-heatmaps}
\end{figure}

\begin{figure}
    \centering
    \includegraphics[width=0.6\columnwidth]{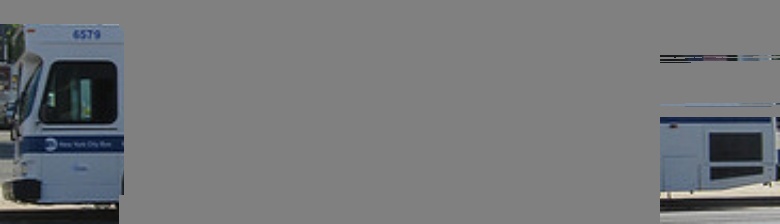}
    \caption{Explanation found by \cet for Fig.~\ref{fig:demo} for the ranking in Fig.~\ref{bus-n50}}
    \label{fig:demo-exp}
\end{figure}

To illustrate how \cet works consider Figure~\ref{fig:demo}, which is classified as \lab{bus} by the compositional net~\cite{kortylewski2020compositional}, even though there is an occlusion in the middle.
\commentout{
\begin{figure}
    \centering
    \includegraphics[width=0.5\columnwidth]{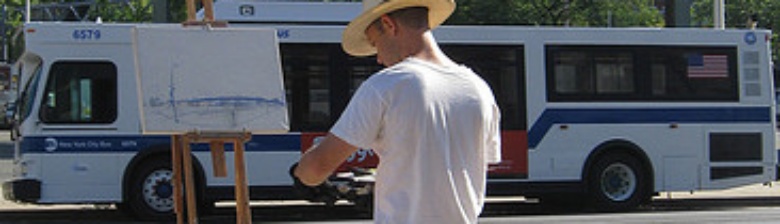}
    \caption{A partially occluded image (from the partial occlusion image data set~\cite{kortylewski2020combining, wang2015unsupervised}) that is classified as \lab{bus} by the compositional net~\cite{kortylewski2020compositional}}
    \label{fig:demo}
\end{figure}
}
Initially, \cet picks up an arbitrary partition of the image with four superpixels, as in Figure~\ref{fig:demo-level0}.
This results in 15 combinations of masking superpixels, which are
given to Alg.~\ref{algo:responsibility} to calculate the responsibility of each superpixel.
The refinement of a superpixel happens in Alg.~\ref{algo:compositonal_responsibility}. As in Figure~\ref{fig:demo-level1},
an initial superpixel is further partitioned into four more fine-grained superpixels. Overall, the refinement done by Alg.~\ref{algo:compositonal_responsibility} for this particular example goes into two levels (Line~7 in Alg.~\ref{algo:compositonal_responsibility}) before the stopping condition is reached (Line~2 in Alg.~\ref{algo:compositonal_responsibility}).
The heat-map in Figure~\ref{fig:demo-heatmaps}~(a) gives the importance
of each pixel hen starting from the single partition in Figure~\ref{fig:demo-level0}.
Though \cet still highlights the important feature in the front of the bus, the result is coarse.
As the number of iterations down by \cet (Alg.~\ref{algo:compositonal_explanation}) increases and a larger number of initial partitions is sampled, the result quickly converges (Figure~\ref{fig:demo-heatmaps}), and the important features are identified successfully, avoiding the occlusion in the input image. The final explanation found by \cet is given as Figure~\ref{fig:demo-exp}.

\commentout{
\begin{figure}
    \centering
    \begin{tabular}{@{}c@{\hspace{0.25cm}}c}
        \includegraphics[width=0.45\linewidth]{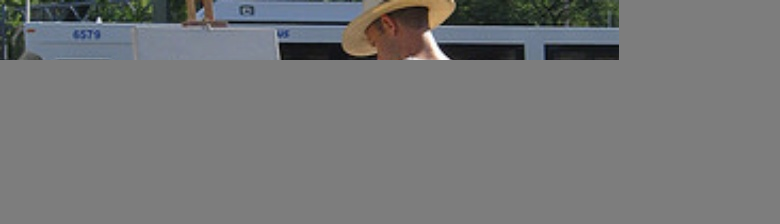} &
        \includegraphics[width=0.45\linewidth]{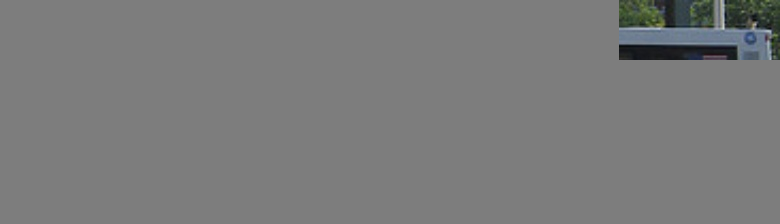}\\
        \includegraphics[width=0.45\linewidth]{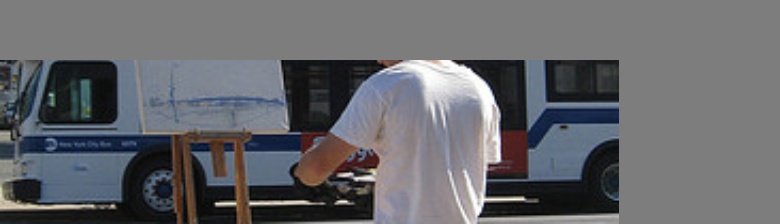} &
        \includegraphics[width=0.45\linewidth]{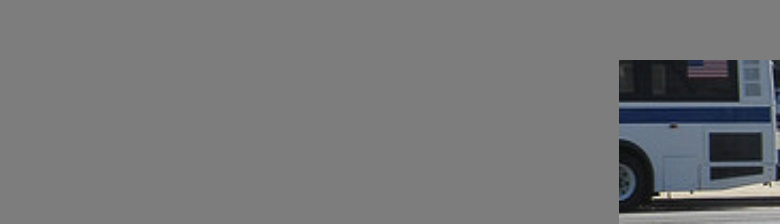} \\
  \end{tabular}
\caption{The initial four superpixels chosen by \cet (Line 3 in Alg.~\ref{algo:compositonal_explanation})}
\label{fig:demo-level0}
\end{figure}

\begin{figure}
    \centering
    \begin{tabular}{@{}c@{\hspace{0.25cm}}c}
        \includegraphics[width=0.45\linewidth]{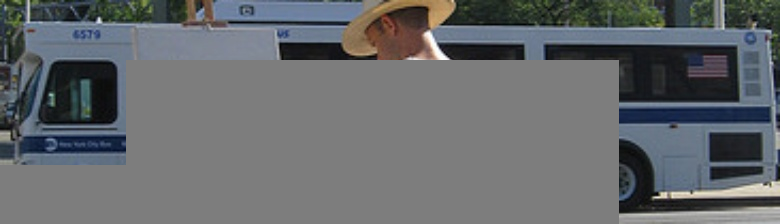} &
        \includegraphics[width=0.45\linewidth]{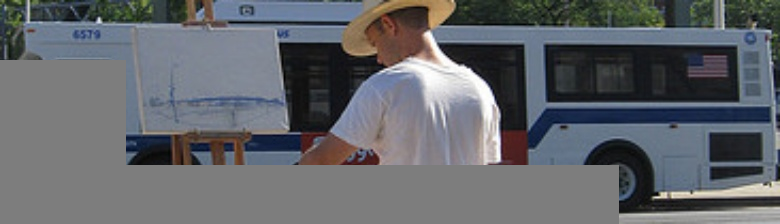}\\
        \includegraphics[width=0.45\linewidth]{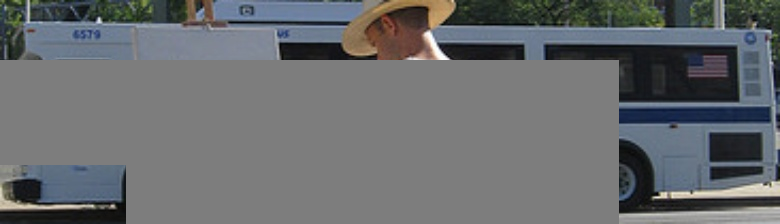} &
        \includegraphics[width=0.45\linewidth]{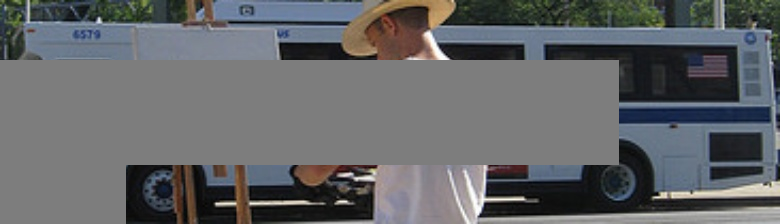} \\
  \end{tabular}
\caption{Further refinement of the (left bottom) superpixel in Figure \ref{fig:demo-level0} (Line 7 in Alg.~\ref{algo:compositonal_responsibility})}
\label{fig:demo-level1}
\end{figure}
}

\subsection{Complexity and comparison with existing work}

We argue that our approach is more efficient than other explanation-generating approaches to image classification.
We first discuss the theoretical complexity of Alg.~\ref{algo:compositonal_explanation}.
Our claims are further supported by the empirical results presented in Sec.~\ref{sec:evaluation}.

The SHAP (SHapley Additive exPlanations) method~\cite{lundberg2017unified} unifies the theory behind several popular methods for explaining AI models. Assuming feature independence, the SHAP approach has runtime complexity $O(2^M + M^3)$, where $M$ is the is the number of input features, bounded by the number of pixels of the input image, and is hence exponential in the size of the input image. 
Even if in practice we rarely observe exponential running time, we stress that there is no guarantee that SHAP finds the important features of the image in reasonable time. The problem becomes even harder when the important features are distributed
in multiple, non-contiguous parts of an input image.

By contrast, our algorithm is based on causal theory and is inherently modular.
Furthermore, the runtime complexity of our algorithm is at most \emph{linear} in the size of image, as 
the following lemma proves.

\begin{lemma}\label{ce-complexity}
The runtime of Alg.~\ref{algo:compositonal_explanation} is $O(2^s n N)$, where $s$ is the size of the partition in each step
(in our setting $s=4$), $n$~is the number of pixels in the original image $x$, and $N$ is the number of initial partitions.
\end{lemma}

\begin{proof}[Proof sketch]
The computation of responsibilities of superpixels in one partition is $O(2^s)$,
as the algorithm examines the effect of mutating each
subset of the superpixels in the current partition. Note that $s$ is a constant independent of the size of the image.
The number of steps is determined by the termination condition on the size of a single superpixel. In our setting,
the algorithm terminates when a single superpixel is $1/10$ of the original image, thus resulting in a constant-time computation.
However, in general, the algorithm can continue down to the level of a single pixel, thus resulting in $n$ pixels in the last step,
hence the factor~$n$. The algorithm performs $N$ iterations, and every iteration uses a different initial partition.
\end{proof}

%% file: evaluation.tex
\section{Evaluation}
\label{sec:evaluation}

\begin{figure}
    \centering
    \subfloat[\lab{West Highland white terrier}]{
    \includegraphics[width=0.22\linewidth]{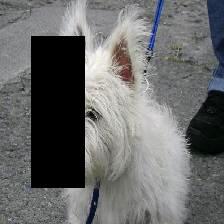}
    \includegraphics[width=0.22\linewidth]{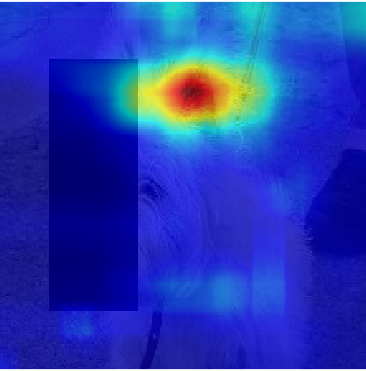}
    }
    \subfloat[\lab{Ocean liner}]{
    \includegraphics[width=0.22\linewidth]{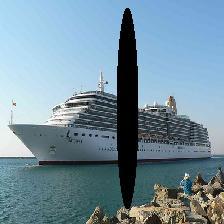}
    \includegraphics[width=0.22\linewidth]{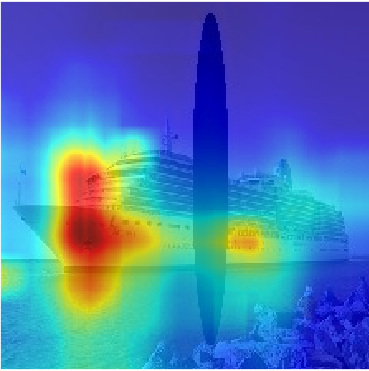}
    }
    \caption{Photo Bombing images and output from \cet}
    \label{fig:photo-bombing-examples}
\end{figure}

\begin{figure*}[t]
\centering
\begin{subfigure}{0.49\textwidth}
    \input{photobombing-intersection}
    \caption{}
    \label{fig:pb-results-intersection}
\end{subfigure}
\begin{subfigure}{0.49\textwidth}
    \input{photobombing-explanation-size}
    \caption{}
    \label{fig:pb-results-size}
\end{subfigure}
\caption{Intersection between the explanation and the occlusion (smaller is better) and the size of the explanation (smaller is better) on the Photo Bombing dataset}
\label{fig:pb-results}
\end{figure*}

\paragraph{Benchmarks and Setup}

We have implemented the proposed compositional explanation approach in the publicly available tool \deepcover\footnote{The experimental data in this section and more results (e.g., for different configurations of the algorithm and explaining misclassifications) can be found at \url{https://www.cprover.org/deepcover/}.}.
In the evaluation, 
we compare \cet with a wide range of explanation tools, including the most recent ones and popular ones like
\deepcoverSBFL~\cite{sun2020explaining},
\extremal~\cite{fong2019understanding},
\rise~\cite{petsiuk2018rise},
\rap~\cite{nam2020relative},
\lrp~\cite{bach2015pixel},
\ig~\cite{sundararajan2017axiomatic}
and
\gbp~\cite{springenberg2015striving}.
We test both the Compositional Net that has been designed for partially occluded images~\cite{kortylewski2020compositional} and standard convolutional models for ImageNet. 
We use three data sets: 
images known to feature partial occlusion~\cite{kortylewski2020combining, wang2015unsupervised}, a ``Photobombing'' data set with added occlusions for which we have  ground truth, and the ``Roaming Panda'' data set, which contains modified ImageNet images without occlusion~\cite{sun2020explaining}. 

There is no single best way to evaluate the quality of an explanation. In this work, we quantify the quality of the explanations with two complementary methods: 1) the explanation size, and 2) the intersection with the planted occlusion part of an image. 
Intuitively, a good explanation should be a part of the original input and it should not intersect much with the occlusion. 

We configure \cet to use four superpixels per partition. The termination conditions for the partition refinement in Alg.~\ref{algo:compositonal_responsibility} are: 1)~the height/width of a superpixel is smaller than $\frac{1}{10}$ of the input image's or 2)~the four superpixels share the same responsibility. The parameter $N$ of Alg.~\ref{algo:compositonal_explanation} is set to~$50$.

\paragraph{MS-COCO images with partial occlusions}
\label{sec:occlu}

\begin{figure}
  \centering
  \subfloat[\lab{train}]{
    \includegraphics[width=0.3\linewidth]{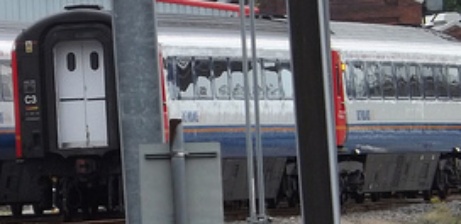}
  }
  \subfloat[\cet]{
    \includegraphics[width=0.3\linewidth]{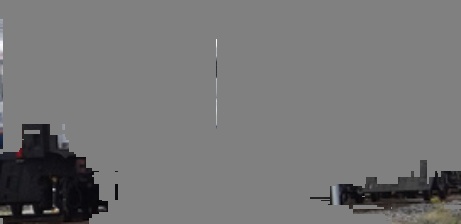}
  }
  \subfloat[\deepcoverSBFL]{
    \includegraphics[width=0.3\linewidth]{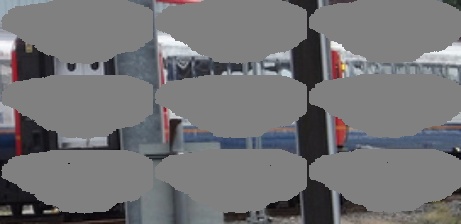}
  }
\caption{Explaining the partially occluded \lab{train}~(a): \cet~(b) vs.~\deepcoverSBFL (c)} 
\label{fig:composition-net-demo}
\end{figure}

In this part of the evaluation, we consider explanations for classifications done by the compositional net~\cite{kortylewski2020compositional} for partially occluded input images from the MS-COCO
dataset \cite{lin2014microsoft}.
Due to the lack of ground truth for this data set (Fig.~\ref{fig:composition-net-demo}),
we use the (normalised) size of the explanation as proxy for quality. 
On average, the explanations from \cet contain less than 13\% of the total pixels and this compares favourably to 37.8\% for \deepcoverSBFL.
\cet's ranking yields 80\% correct classification results when using the Compositional Net on only 20\% of the pixels of the input image. This is more than a 20\% improvement over \deepcoverSBFL.

\commentout{
\begin{figure}
    \centering
    \includegraphics[width=0.95\columnwidth]{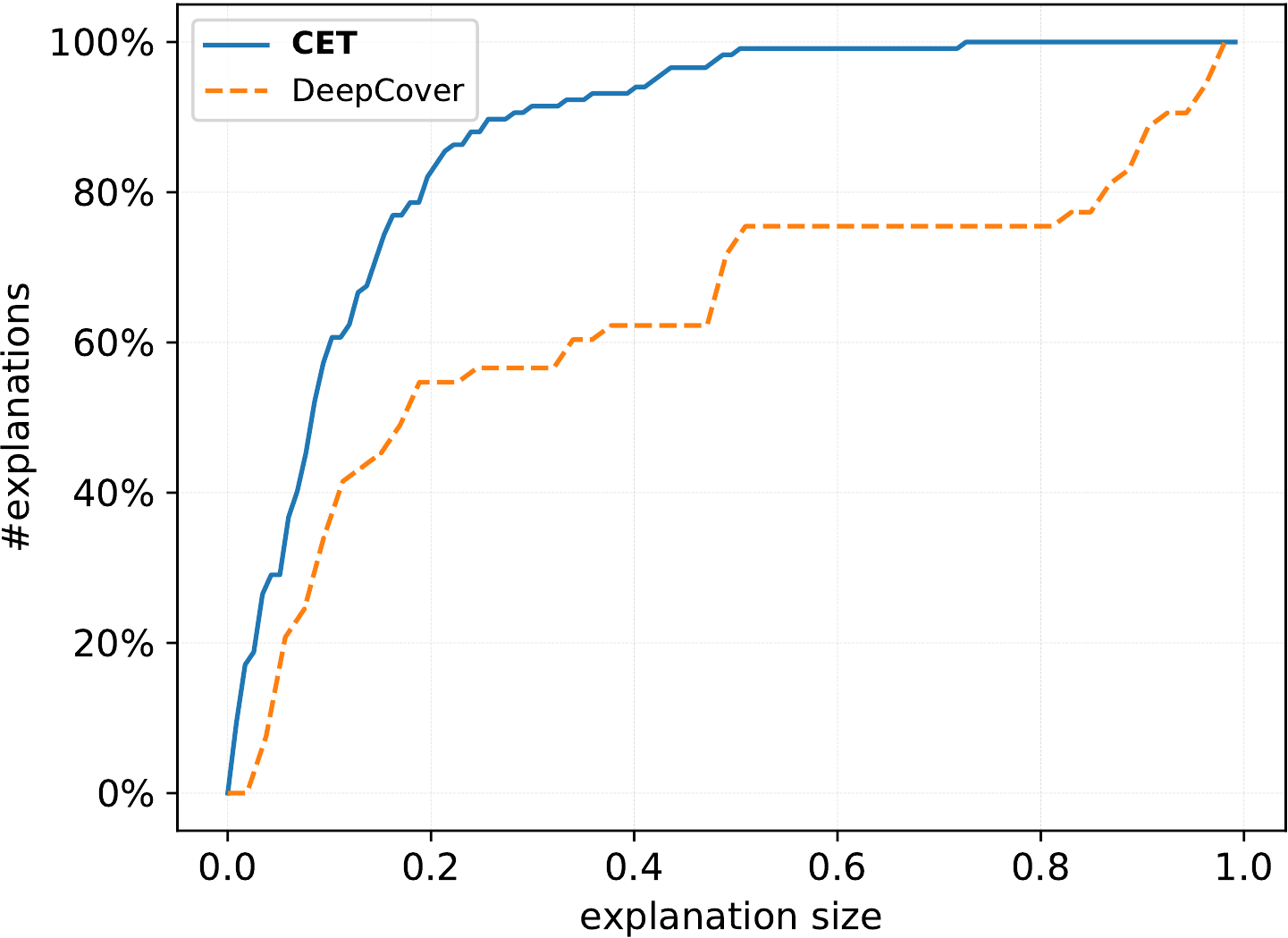}
    \caption{Sizes of explanations for the compositional net}
    \label{fig:results-compositional-net}
\end{figure}
}
\paragraph{``Photo Bombing'' images}
\label{sec:photo-bombing}

Similarly to the images with partial occlusion used in Section~\ref{sec:occlu}, we create an image set named ``Photo Bombing''. We plant occlusions (aka ``photobombers'') into ImageNet images and we record these occlusion pixels so that we can measure the intersection of explanations with the occluded pixels. Examples (and the corresponding \cet explanations) from the Photo Bombing data set are given in Fig.~\ref{fig:photo-bombing-examples}. 

Fig.~\ref{fig:pb-results} gives the results on the photobombing images. The $y$-axis is the retained accuracy of the classifier on the explanation part of the image (with the rest masked). We observe that more than $60\%$ of the \cet explanations do not 
have any overlap with the artificially planted occlusions, which is $20\%$ better than than the second best 
tool \rise. The explanation size from \cet is also consistently smaller than other tools, as shown in Fig.~\ref{fig:pb-results-size}. Interestingly, even though \cet constructs smaller explanations than \rise, its explanations have a higher overlap with the occlusions, thus illustrating the necessity of using
more than one quality measure for assessing the quality of explanations. 

\paragraph{Masking colors}

When using \cet (or any explanation algorithm that uses input perturbation), we need a masking color for removing parts of the input image. We~argue empirically that the choice of masking color has little to no impact on the performance of our algorithm. To this end, we~have run our experiments with a number of masking colors, ranging from black $(0,0,0)$ to white $(255,255,255)$. We measured the mean intersection between \cet's explanations and the occlusions for each masking color. We~have observed that the influence of the masking color is less than~$4\%$. The full results for this experiment are on the project website.

\paragraph{``Roaming Panda'' images}
To determine the performance of \cet on images without occlusion, we have performed an experiment on the ``Roaming Panda'' dataset~\cite{sun2020explaining}. The classifier detects the Panda, which is superimposed and therefore never occluded in this dataset.
We have compared \cet with the other explanation tools by measuring the intersection of union (IoU) between the explanation and the panda part (the larger the better). \cet achieved a (very close) third place, demonstrating that its performance on non-occluded images is comparable to the best explanation tools. The full results are available on the project website.

%% file: photobombing-intersection.tex
\begin{tikzpicture}

\begin{axis}[
  yticklabel={$\pgfmathprintnumber{\tick}\%$},
  xlabel={intersection},
  ylabel={retained accuracy},
  legend pos=south east,
  legend style={font=\scriptsize},
  legend cell align={left},
  legend columns=2,
  tick label style={font=\scriptsize},
  grid=both,
  grid style={line width=.1pt, draw=gray!10}]

\pgfplotsset{
    legend entry/.initial=,
    every axis plot post/.code={%
        \pgfkeysgetvalue{/pgfplots/legend entry}\tempValue
        \ifx\tempValue\empty
            \pgfkeysalso{/pgfplots/forget plot}%
        \else
            \expandafter\addlegendentry\expandafter{\tempValue}%
        \fi
    },
}

\pgfplotsset{y filter/.code={\pgfmathparse{#1*100}\pgfmathresult}}

\addplot[mark=none, ta2skyblue, legend entry=\tool{\textbf{dc-causal}}, thick] table {photobombing-intersection/causal.txt};
\addplot[mark=none, ta3chameleon, legend entry=\rise, dotted, thick] table {photobombing-intersection/rise.txt};
\addplot[mark=none, ta2orange, legend entry=\tool{dc-sbfl}, dashed, thick] table {photobombing-intersection/sbfl.txt};
\addplot[mark=none, ta2plum, legend entry=\rap, dashed, thick] table {photobombing-intersection/rap.txt};
\addplot[mark=none, pink!50!red, legend entry=\gbp, dashed, thick] table {photobombing-intersection/gbp.txt};
\addplot[mark=none, brown, legend entry=\lrp, dotted, thick] table {photobombing-intersection/lrp.txt};
\addplot[mark=none, ta2scarletred, legend entry=\extremal, thick] table {photobombing-intersection/ext.txt};
\addplot[mark=none, ta2butter, legend entry=\ig, thick] table {photobombing-intersection/ig.txt};

\end{axis}
\end{tikzpicture}
\vspace*{-0.15cm}

%% file: photobombing-explanation-size.tex
\begin{tikzpicture}

\begin{axis}[
  yticklabel={$\pgfmathprintnumber{\tick}\%$},
  xlabel={explanation size},
  ylabel={retained accuracy},
  legend pos=south east,
  legend style={font=\scriptsize},
  legend cell align={left},
  legend columns=2,
  tick label style={font=\scriptsize},
  grid=both,
  grid style={line width=.1pt, draw=gray!10}]

\pgfplotsset{
    legend entry/.initial=,
    every axis plot post/.code={%
        \pgfkeysgetvalue{/pgfplots/legend entry}\tempValue
        \ifx\tempValue\empty
            \pgfkeysalso{/pgfplots/forget plot}%
        \else
            \expandafter\addlegendentry\expandafter{\tempValue}%
        \fi
    },
}

\pgfplotsset{y filter/.code={\pgfmathparse{#1*100}\pgfmathresult}}

\addplot[mark=none, ta2skyblue, legend entry=\tool{\textbf{dc-causal}}, thick] table {photobombing-explanation-size/causal.txt};
\addplot[mark=none, ta3chameleon, legend entry=\rise, dotted, thick] table {photobombing-explanation-size/rise.txt};
\addplot[mark=none, ta2orange, legend entry=\tool{dc-sbfl}, dashed, thick] table {photobombing-explanation-size/sbfl.txt};
\addplot[mark=none, ta2plum, legend entry=\rap, dashed, thick] table {photobombing-explanation-size/rap.txt};
\addplot[mark=none, pink!50!red, legend entry=\gbp, dashed, thick] table {photobombing-explanation-size/gbp.txt};
\addplot[mark=none, brown, legend entry=\lrp, dotted, thick] table {photobombing-explanation-size/lrp.txt};
\addplot[mark=none, ta2scarletred, legend entry=\extremal, thick] table {photobombing-explanation-size/ext.txt};
\addplot[mark=none, ta2butter, legend entry=\ig, thick] table {photobombing-explanation-size/ig.txt};

\end{axis}
\end{tikzpicture}
\vspace*{-0.15cm}

%% file: conclusions.tex
\section{Conclusions}
\label{sec:conclusions}

We propose a causal approach for explaining the output of image classifiers. Based on the definitions
in~\cite{HP05}, an explanation is a minimal subset of pixels of the image that is sufficient
for the classification. As the exact computation is intractable~\cite{CH04}, we describe a modular and parallelizable
algorithm for computing an approximation to the explanation using a \emph{causal ranking} of parts of the
input image. Our experiments demonstrate that \cet produces accurate results for partially occluded images, which pose a challenge to other explanation tools.

\section*{Acknowledgments}
The authors acknowledge funding from the UKRI Trustworthy Autonomous Systems Hub (EP/V00784X/1) and the UKRI Strategic Priorities Fund to the UKRI Research Node on Trustworthy Autonomous Systems Governance and Regulation (EP/V026607/1).